%% file: main.tex
\documentclass{article}
\usepackage{spconf,amsmath,graphicx,amssymb, booktabs}
\usepackage{bbm}
\usepackage{multirow}
\usepackage{hyperref}
\usepackage{textcomp}
\usepackage{xcolor}
\usepackage[inline]{enumitem}
\usepackage{siunitx}
\usepackage[nameinlink]{cleveref}
\newcommand{\tworow}[1]{\multirow{2}{*}{\centering #1}}

\title{Lego-Features: Exporting modular encoder features \\ for streaming and deliberation ASR}
\name{Rami Botros, Rohit Prabhavalkar, Johan Schalkwyk, Ciprian Chelba, Tara N. Sainath, Françoise Beaufays }
\address{Google LLC, USA \\
\fontsize{9}{9}\selectfont\ttfamily\upshape
ramibotros@google.com}
\begin{document}
\ninept
\maketitle

\input{abstract}

\def\arraystretch{1.1}
\input{introduction}

\input{modeling}

\input{experimental_settings}

\input{results}
\input{conclusion}

\bibliographystyle{IEEEbib}
\bibliography{refs}

\end{document}

%% file: abstract.tex
\begin{abstract}
In end-to-end (E2E) speech recognition models, a representational tight-coupling inevitably emerges between the encoder and the decoder. We build upon recent work that has begun to explore building encoders with modular encoded representations, such that encoders and decoders from different models can be stitched together in a zero-shot manner without further fine-tuning. While previous research only addresses full-context speech models, we explore the problem in a streaming setting as well. Our framework builds on top of existing encoded representations, converting them to modular features, dubbed as \emph{Lego-Features}, without modifying the pre-trained model. The features remain interchangeable when the model is retrained with distinct initializations. Though sparse, we show that the Lego-Features are powerful when tested with RNN-T or LAS decoders, maintaining high-quality downstream performance. They are also rich enough to represent the first-pass prediction during two-pass deliberation. In this scenario, they outperform the N-best hypotheses, since they do not need to be supplemented with acoustic features to deliver the best results. Moreover, generating the Lego-Features does not require beam search or auto-regressive computation. Overall, they present a modular, powerful and cheap alternative to the standard encoder output, as well as the N-best hypotheses.
\end{abstract}
\begin{keywords}
modular, representations, zero-shot stitching
\end{keywords}

%% file: introduction.tex
\section{Introduction} \label{sec:intro}

E2E speech recognition models, which combine acoustic, pronunciation and language models from conventional systems \cite{Golan16} into one neural network, have become widely used, especially for on-device applications \cite{bo21system,Ryan19,CC18,KimHoriWatanabe17,JinyuLi2019,Zeyer2020}. Since they are much smaller than conventional models, and their inference speed is often much faster \cite{bo21system,Ryan19,sainath2021cascadedlm,sainath2020streaming}, they work well for various streaming applications. They typically use an encoder-decoder architecture \cite{RohitSeq17}. Like most deep neural networks, the whole architecture is usually trained end to end. The encoder implicitly learns to serve the subsequent decoder layers, and thus conversely, the decoder is thoroughly oriented towards inputs coming from the specific encoder that it has been trained with. Therefore, encoders and decoders from different models or training runs, are generally not interchangeable without further E2E training.

This tight coupling between both components stands in the way of a flexible, modular architecture. Speech encoders that have been trained on high-resource ASR data can serve as foundation models for other tasks like sentiment analysis \cite{lu2020speech} or low-resource translation \cite{bansal2018pre}, to name a few. However, this presents a challenge if a shared encoder representation is used for multiple downstream tasks: When the ASR encoder is retrained, all downstream models must be retrained as well. Hence, it would be more practical if each component can be developed and updated independently. To that end, we present a method for building modular speech encoder features, where different versions of the encoder can be plugged into the decoder in a zero-shot stitching manner without fine-tuning.

Our method works by building on top of an existing base encoder, which is kept frozen. We adapt the Beam-Convolution scheme described in \cite{dalmia2019enforcing} to train streaming modular encoded representations, which we call Lego-Features. To produce them, the original (fixed) continuous encoded features pass through a few extra trainable ``Exporter'' layers, then through a CTC decoder, which is trained with an auxiliary CTC loss. Lego-Features are defined as the sorted top $K$ CTC logit indices at every frame, see \Cref{fig:lego_encoder}. The logits operate over a discrete space (here: wordpiece vocabulary) and are grounded in the transcript text, which is why they tend to be modular. Overall, the traditional encoder features are forced through a tight discretizing bottleneck, which protects downstream models from coupling themselves to fine details in the encoded representation. Downstream consumers of Lego-Features need to first re-embed them, since they come in as sparse indices.

\cite{dalmia2019enforcing, dalmia2022legonn} have shown how this tight bottleneck still produces a powerful representation which is sufficiently informative for downstream ASR decoders. They also perform a ``modularity test'': The downstream decoder is kept constant, but gets input with a new version of the encoded representation, which is obtained by retraining the encoder from scratch using a different initialization. The switch is done in a zero-shot manner without any extra fine-tuning. Traditional continuous encoded features categorically fail the modularity test, bringing the downstream performance to nearly 100\% WER, which is what motivates this new type of encoded representation. We build on the original works with a few novel contributions:

\begin{enumerate}[align=right,itemindent=2em,labelsep=4pt,labelwidth=1em,leftmargin=0pt,nosep,label={\arabic*)}]
\item  We find that training the modular encoder from scratch under the CTC loss is insufficient for producing the best performance. Instead, our recipe pre-trains some base encoder layers with RNN-T loss and keeps them frozen. Next, we just train the extra Exporter layers with the auxiliary CTC loss. This solution is also practical since it enables researchers to cheaply export modular features without having to modify their original system. Thus, the quality, latency and efficiency of the base model are all maintained.

\item We adapt the design to a streaming setting for the first time. Unlike the original work \cite{dalmia2019enforcing, dalmia2022legonn}, our encoder layers attention have limited left and right context windows, and the produced Lego-Features are successfully paired with a streaming-friendly RNN-T decoder. The streaming architecture still exhibits strong downstream ASR quality and passes the modularity test. By plugging the same fixed set of Lego-Features into causal as well as non-causal decoders, our work adds further evidence to their modularity and interoperability.

\item Rather than merely looking at the Lego-Features as an encoded representation, we also study them as an alternative to the N-best hypotheses within two-pass systems. We provide new comparisons against the N-best in terms of speed, accuracy and modularity. To this end, the Lego-Features are used as a first-pass output within the deliberation framework \cite{hu2020deliberation}. This achieves good post-deliberation WER performance, which is shown to be on-par with a baseline that performs deliberation on \mbox{1st-pass} \mbox{RNN-T} \mbox{N-best} hypotheses + audio features. The Lego-Features demonstrate success in the modularity test here as well. On the other hand, we find that the N-best hypothesis text does not pass the modularity test, i.e. a new N-best from a second model would confuse the deliberation decoder from the first, which is a novel observation. Moreover, the Lego-Features are cheaper to produce than the N-best, since they require no beam-search or auto-regressive decoding, but are generated via a simple projection at every frame.
\end{enumerate}
Other works have attempted to present generic methods for zero-shot stitching between layers. In \cite{moschella2022relative}, this is achieved by learning representations relative to data-dependent anchors. In contrast, the method presented here does not need to choose anchor samples and leverages the existence of ground-truth speech transcripts instead. Another general approach, presented in \cite{gygli2021towards}, uses self-supervised objectives designed to encourage compatibility of different layer outputs. It is an open question whether the cited methods can deal with long sequences, whereas the CTC loss used here is a natural choice that works well with ASR and gives interpretable outputs.

Further, some research has already experimented with deliberation on top of CTC outputs to save the cost of first-pass decoding \cite{chi2020align,wang2022deliberation,wang2022streaming}. This includes the Align-refine approach, which iteratively improves on the first-pass output. Those works tend to focus on optimizing the size and speed of the first-pass model, whereas our focus is mainly on modularity. Nevertheless, since we build on base encoder layers that have been pre-trained with the RNN-T loss, we find our CTC outputs to have high quality, which removes the need for audio attention that is used in other deliberation models. Hence, this work also introduces some speed gains to deliberation, without using the iterative Align-refine approach.

On the whole, with one simple representations, we get a compelling cheap, streaming-friendly, as well as modular, alternative to both the continuous encoding vector and the N-best hypotheses, without any loss in quality.

%% file: modeling.tex
\section{Modeling} \label{sec:modeling}

Our framework is trained in three separate stages described below.

\subsection{Base Model}
\label{subsec:model_base_model}

\begin{figure}[t]

\center
  \includegraphics[width=0.5 \textwidth]{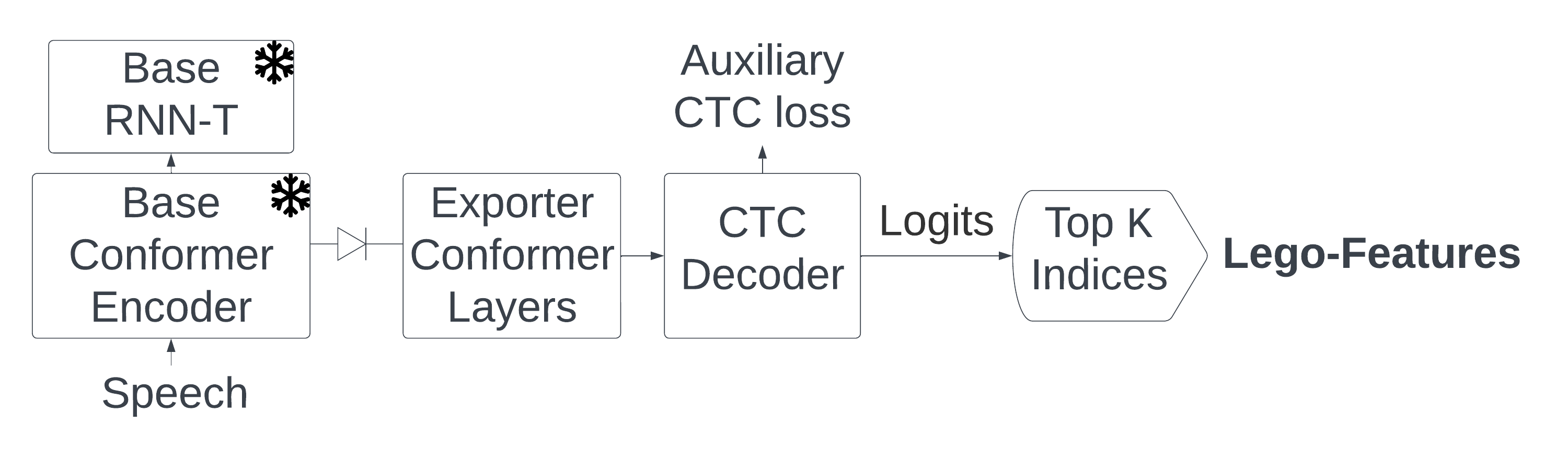}
  \caption{Modular Encoder. Lego-Features are exported from frozen base encoder by training extra layers with an auxiliary CTC loss.}
  \label{fig:lego_encoder}
\end{figure}

We start off from a pre-trained end-to-end system that follows the cascade architecture in \cite{sainath2022improving}: The base encoder comprises 3 convolution layers, then 14 Conformer \cite{gulati2020conformer} blocks: 4 causal ones, followed by 5 blocks that process 180 milliseconds of right-context each, then 5 more causal ones. This base encoder is pre-trained using the RNN-T loss on the same training set. For the modularization steps below, the pre-trained \mbox{RNN-T} decoder layers will be discarded, and the base encoder is kept frozen. This recipe allows us to keep the existing pre-trained model unchanged while exporting modular features.

\subsection{Exporting Lego-Features}
\label{subsec:model_exporting}

\Cref{fig:lego_encoder} shows how the modular encoder is trained on top of a frozen base model. The Exporter layers comprise further Conformer blocks with 180ms look-ahead context. The CTC decoder \cite{Graves06} amounts to a single  projection layer to compute the frame-level posterior over the output vocabulary. Our work uses wordpiece output tokens, but further research can explore using phonemes or graphemes instead. The depicted CTC loss is applied to those logits and is what trains the Exporter layers. Finally, the Lego-Features are computed by extracting the sorted top-$K$ indices of the CTC logits, giving $K$ integers at every frame. Note that this is performed on the logit vector directly, without requiring any actual decoding algorithm like beam-search.

\subsection{Downstream Models}
\label{subsec:model_downstream}

\begin{figure}[t]

\center
  \includegraphics[width=0.5 \textwidth]{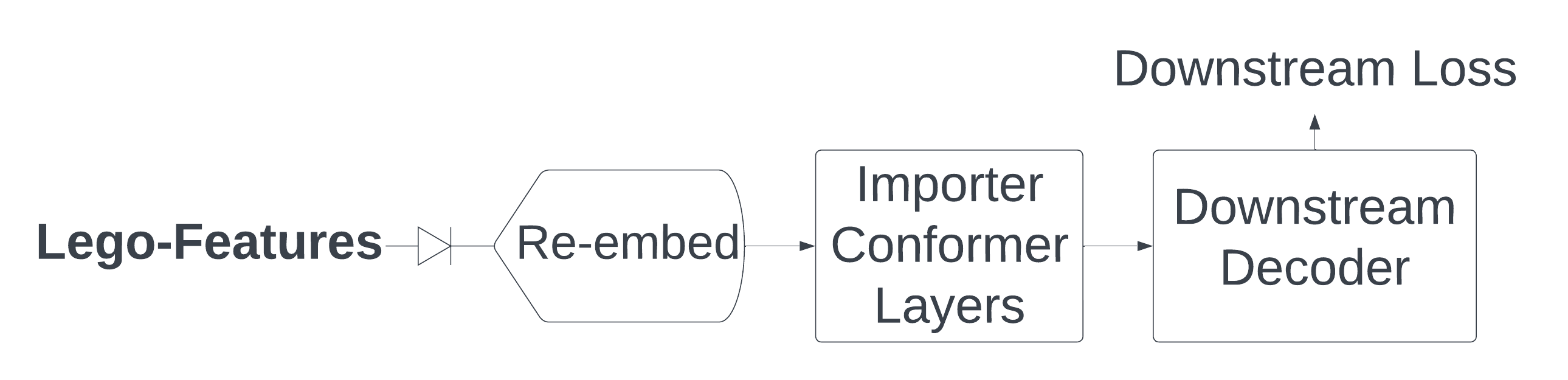}
  \caption{Downstream models embed and process the fixed Lego-features before passing them to a downstream decoder.}
  \label{fig:downstream_decoder}
\end{figure}

\Cref{fig:downstream_decoder} illustrates how downstream models generally consume the Lego-Features, which come in as sparse indices. The downstream consumer does not receive extra information about how the indices map to wordpiece tokens, and hence starts by embedding them. An Importer module, once again consisting of 180ms look-ahead Conformer blocks, prepares the embeddings for the downstream decoder. \cite{dalmia2019enforcing, dalmia2022legonn} use 1D convolution + multi-headed attention in place of the Importer, but our early experiments show that Conformer blocks improve over this original stack. Note that the Lego-Features themselves are kept constant during downstream training. We experiment with two types of ASR decoders as examples for downstream tasks, which are used with the same fixed set of Lego-Features.

\subsubsection{Downstream RNN-T Decoder}
\label{subsubsec:model_downstream_rnnt}
The first downstream model uses an RNN-T decoder, which tends to serve real-time applications well, since it processes the input frames in a streaming fashion as they become available and starts outputting text tokens after a short delay \cite{Ryan19, Graves12}. We adopt the same RNN-T decoder layer architecture from the base model (\Cref{subsec:model_base_model}) but use it as a simulated downstream task, as the decoder in \Cref{fig:downstream_decoder}, to see if the bottlenecked Lego-Features are as informative as the continuous base encoded tensor. 

\subsubsection{Downstream LAS decoder / Deliberation}
\label{subsubsec:model_downstream_las}

\begin{figure*}[t]
  \center
  \includegraphics[width=\textwidth]{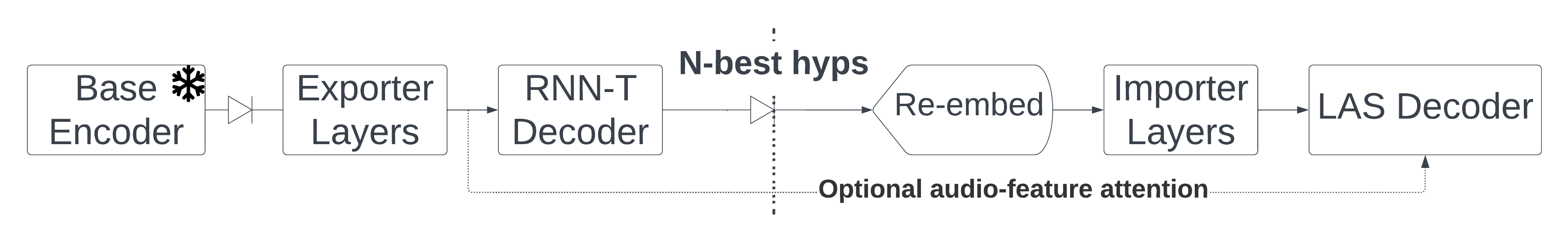}
  \caption{Baseline deliberation on N-best RNN-T hyps. The LAS decoder attends to embedded text and optionally to the pre-RNN-T audio features. Modularity test boundary shown as the dotted line in the middle.}
  \label{fig:rnnt_delib_full}
\end{figure*}

As a second downstream ASR decoder in \Cref{fig:downstream_decoder}, we experiment with a full-context Listen-Attend-and-Spell (LAS) decoder \cite{Chan15}, which can achieve higher quality by attending to all input frames.

A fitting baseline to this experiment is second-pass deliberation ASR \cite{hu2020deliberation}. Typically, a deliberation system generates first-pass hypotheses using a fast decoder, like RNN-T, then embeds its N-best hyps and attends to them with a second-pass full-context LAS decoder. We have therefore constructed a comparable deliberation baseline model shown in \Cref{fig:rnnt_delib_full}. This model is analogous to our full pipeline, i.e. Figures \ref{fig:lego_encoder} \& \ref{fig:downstream_decoder} put together, and is designed to have a similar total model size and encoder latency. It starts with the same frozen base encoder, then trains a first-pass RNN-T decoder to obtain the N-best hyps, which stands to be compared to the Lego-Features in terms of informativeness and modularity. \Cref{fig:rnnt_delib_full} also ends with an LAS decoder, except this one can optionally attend to the continuous encoder features as well, as is done in previous deliberation work \cite{hu2020deliberation}. Gradients do not flow back through embedded N-best.

%% file: experimental_settings.tex
\section{Experimental settings} \label{sec:experiments}
\subsection{CTC Logit Evaluation}
\label{subsec:exp_settings_ctc_logit_eval}
An interesting aspect of the Lego-Features encoder is that one can evaluate its quality directly before providing the features to any downstream tasks. This is done via a preliminary experiment where we directly decode from the full set of the CTC-trained logits (before the top-$K$ operation in \Cref{fig:lego_encoder}) using beam search or greedy decoding. The decoding algorithm used for this evaluation is tangential to how the Lego-Features are produced, since those are only extracted as the top-$K$ logit ranks without decoding actual transcripts. Yet this direct evaluation can inform us about the general quality of the CTC-trained logits, from which the Lego-Features are produced.

\subsection{WER and Modularity Test}
The downstream ASR decoders trained on the Lego-Features (\Cref{subsec:model_downstream}) are then evaluated and a modularity test is performed. The aim of the test is to check if two different versions of the encoded features are interchangeable. We test that by keeping the downstream model fixed, but feeding it with a new version of the encoded features, which we get from another training run. The second training is done from scratch with a new initialization. We compare the WER performance of the decode before and after the switch, denoted as ``Normal $\to$ Mod. Test WER'' in our tables. For the Lego-Features, we retrain the encoder in \Cref{fig:lego_encoder}, where the base frozen encoder is also replaced with a second version from a retrained base. As a baseline, we also test the modularity of the base model itself, where we simply train the base encoder + decoder a second time end-to-end and get the retrained encoder from there.

\subsection{Architectural Details}

Our base architecture follows \cite{sainath2022improving}: All Conformer layers \cite{gulati2020conformer} are 512-dim, use 8-headed self-attention and a convolution kernel size of 15. We train on a 128D log-mel feature frontend with a 16-D one-hot domain-id vector appended to it, see \cite{Arun19}.

Our models work with 4,096 word pieces~\cite{Schuster2012}. The RNN-T decoder comprises a prediction network and a joint network with a single 640-dim FF layer. The embedding prediction network \cite{Rami21}, uses an embedding dimension of 320, and has 9M parameters. For the deliberation decoder, we use a 2-layer LSTM similar to \cite{hu2020deliberation}, where each layer has 1536 hidden units followed by 384-dim projection. We do not use external LMs.

\subsection{Datasets}

As discussed in \cite{sainath20streaming}, all E2E models are trained on multidomain audio-text pairs \cite{Arun19}. All datasets obtain their labels in a semi-supervised fashion, using larger teacher models trained on in-domain data to provide pseudo labels \cite{Seong22,liao2013large}. Data was handled in accordance to Google AI principles \cite{google_ai_principles}. To further increase data diversity, multi-condition training (MTR) ~\cite{kim2017mtr}, random data down-sampling to 8kHz \cite{Li12} and SpecAug \cite{Park2019} are also used. Noisy data is generated at signal-noise-ratio (SNR) from 0 to 30~dB, with an average SNR of 12~dB, and with T60 times ranging from 0 to 900ms, averaging 500ms. Noise segments are sampled from YouTube and daily life noisy environmental recordings. Both 8~kHz and 16~kHz versions of the data are generated, each with equal probability, to make the model robust to varying sample rates. 

The \emph{Voice-Search} test set has 10K Voice Search utterances with an average length of 5.5 seconds. They are anonymized, hand-transcribed, and are representative of Google's Voice Search traffic.

%% file: results.tex
\section{Experimental Results} \label{sec:results}

\subsection{Preliminary CTC Decoder Evaluation}

\input{tables/tbl_1_ctc_wer}

As explained in \Cref{subsec:exp_settings_ctc_logit_eval}, the CTC decoder in \Cref{fig:lego_encoder} can be evaluated directly. \Cref{tbl:ctc_wer} shows two settings for the Exporter layers and their corresponding CTC WER performance. The right-context length indicates the extra duration of future context attended to by the Exporter, noting that the base encoder already sees a future context of 900ms. In both cases, greedy decoding performs close to beam search, which tracks 16 hypotheses in its beam. For all the downstream experiments below, we use the better setup with 3 blocks for the Exporter, and apply the same design to the Importer.

\subsection{Base RNN-T vs. Downstream RNN-T}

Our first downstream setting works with an RNN-T decoder (\Cref{subsubsec:model_downstream_rnnt}).
\Cref{tbl:downstream_rnnt} demonstrates how the Lego-Features bottleneck still produces a rich encoding that the downstream Importer and RNN-T use well. We export $K=12$ Lego-Features per frame and the downstream re-embeds each into $32$ dimensions. Preliminary experiments, omitted here for brevity, indicate that varying these values does not affect downstream WER performance significantly.
The Base case in the table is simply the frozen base model on the left of \Cref{fig:lego_encoder}, in which case the modularity test connects a new base encoder (from another training run) to the same frozen base RNN-T. The modularity test fails for the base case, yet passes for the Lego-Features. Both models involve different sizes and latencies, so a direct WER contest between them is not the main concern. Rather, the goal is to show that the Lego-Features bottleneck does not degrade performance while enabling modularity.

To test robustness across changing domains, we also supply the same Lego-Features used above to a downstream RNN-T model that is trained on Librispeech data instead. The modularity test results are shown in \Cref{tbl:downstream_librispeech_rnnt} and only cause less than 4\% relative WER decline.

\input{tables/tbl_2_downstream_rnnt}

\subsection{Deliberation on N-Best vs. Lego-Features}
\Cref{tbl:downstream_deliberation} compares the LAS deliberation scenarios described in \Cref{subsubsec:model_downstream_las}, where the Lego-Features are compared to an N-best as a first-pass output. Dropping the audio connection significantly degrades performance in the N-best case, which is consistent with previous findings \cite{hu2020deliberation}. The Lego-Features seem to preserve more information in the encoding, and thus do not need the audio connection. They are significantly better than N-best text, and are only off by 0.1 in absolute WER from N-best + audio.

The modularity test causes no performance decline for the Lego-Features, but does not work well in the N-best case; even the text-only case degrades by 17\% relative WER. This somewhat unexpected result might be a symptom of label bias, which RNN-T suffers from because of local normalization \cite{hannun2020label,yan2022ctc}, but the CTC decoder avoids with its conditional independence assumption. Hence, two separately-trained RNN-T first-pass models might exhibit different biases in their N-bests, leading to this result.

\input{tables/tbl_3_downstream_libir_rnnt}

\subsubsection{Speed Comparison}

\label{subsec:delib_speeds}
\Cref{tbl:downstream_deliberation} notes a difference in the input shapes to the Importers across the different types of first-pass models, after re-embedding in \Cref{fig:downstream_decoder} \& \ref{fig:rnnt_delib_full}. Here, $E_1$ and $E_2$ are the respective embedding dimensions, $n$ is the \mbox{RNN-T's} beam width and $U$ is the number of text tokens produced by it. $K$ is the number of logit indices in the Lego-Features and $T$ is their sequence length (=number of encoded frames). Note how the N-best's embedding expands the output sequence length,
since it stacks the $N$ hypothesis sequentially while keeping the sentence structures intact, in order to attend to this order information during second-pass decoding. Since the LegoFeatures describe per-frame logit ranks without serializing them into sentences, we forgo this expansion and concatenate the embeddings within the depth dimension at each frame instead. This saves on computational cost, since the \#GFLOPs used by LAS is proportional to the sequence length it is attending to. While $U$ can change from one utterance to the other, the embedded matrices have to padded to maximum length when working with hardware accelerators. Our system uses $n=8$, $U=120$, $E_1=384$, $T=343$, $K=12$, and $E_2=32$. This makes the depth dimension equal, but LegoFeatures' sequence length is $64\%$ smaller than the N-best's.

Another important computational benefit of deliberating on LegoFeatures is that we can obtain them without performing a beam-search procedure. It is hence  possible to compute them for long utterances with high parallelization, only limited by the number of TPU cores available. Generating the N-best, on the other hand, requires sequential auto-regressive processing. For instance, benchmarking this sequential path in the RNN-T (using an in-house server TPU and the above dimensions) gives \mbox{$1.8$ ms} per output token, or $216$ ms per utterance in the padded worst case, which does become the bottleneck after the other layers are parallelized.

\input{tables/tbl_4_downstream_deliberation}

%% file: tables/tbl_1_ctc_wer.tex
\begin{table}[h]
    \centering
    \begin{tabular}{ccc|cc} \toprule
    \multicolumn{3}{c|}{Exporter Properties} & \multicolumn{2}{c}{CTC Test WER} \\
    \tworow{\# Blocks} & \tworow{Size} & Right & \tworow{Greedy} & Beam-search \\
    &   & Context  &   & (Oracle) \\ \midrule
    1 & 10M & +180 ms & 5.9\% & 5.8\% (2.8\%) \\
    3 & 30M & +540 ms & \textbf{5.5\%} & \textbf{5.3\% (2.7\%)} \\\bottomrule
    \end{tabular}
    \caption{CTC Voice-Search WER for different Exporter setups}
    \label{tbl:ctc_wer}
\end{table}


%% file: tables/tbl_2_downstream_rnnt.tex

\begin{table}[t]
    \centering
    \begin{tabular}{lcc|r@{ $\to$ }l} \toprule
    \multicolumn{3}{c|}{Encoder}  & \multicolumn{2}{c}{RNN-T  WER} \\
    Type & Size & Right-Context &  Normal & Mod. Test \\ \midrule
    Base & 146M & \phantom{0}900 ms & 6.4\% & 99\% \\ 
    Modularized & 207M & 1440 ms  & 5.6\% & \textbf{5.6\%} \\\bottomrule
    \end{tabular}
    \caption{Downstream RNN-T Test WER with Modularity Test. The base encoder is from the original pre-trained model. }
    \label{tbl:downstream_rnnt}
\end{table}

%% file: tables/tbl_3_downstream_libir_rnnt.tex
\begin{table}[b]
    \centering
    \begin{tabular}{r@{ $\to$ }l|r@{ $\to$ }l} \toprule
    \multicolumn{2}{c|}{Dev-Clean WER}  & \multicolumn{2}{c}{Test-Other WER} \\
    Normal & Mod. Test & Normal & Mod. Test \\ \midrule
    4.9 & \textbf{5.1} & 10.0 & \textbf{10.3} \\\bottomrule
    \end{tabular}
    \caption{Modularity tests if downstream is trained on Librispeech}
    \label{tbl:downstream_librispeech_rnnt}
\end{table}

%% file: tables/tbl_4_downstream_deliberation.tex

\begin{table}[t]
    \centering
    \begin{tabular}{l|c|c|r@{ $\to$ }l} \toprule
    \tworow{First Pass} & Embedded & Attend & \multicolumn{2}{c}{Downstream WER} \\
    & Shape & Audio & Normal & Mod. Test \\ \midrule
     \multirow{2}{*}{\centering RNN-T $N$-best}  & \tworow{ $\left[ N \cdot U, E_1 \right] $} & No & 5.4\% & \phantom{0}6.3\%  \\ 
      &                                            &  Yes & 5.0\% & 14.3\%  \\ \hline
     Lego-Features & $\left[ T, K \cdot E_2 \right] $  & No & 5.1\% & \phantom{0}\textbf{5.1\%}  \\ \bottomrule
     \end{tabular}
    \caption{Deliberation WER and Modularity Tests. Embedded Shapes discussed in \Cref{subsec:delib_speeds}}
    \label{tbl:downstream_deliberation}
\end{table}

%% file: conclusion.tex
\section{Conclusions and Future Work}
In this paper, we describe a simple recipe for exporting streaming-friendly modular encoded representations and successfully test them with RNN-T and LAS decoders. Overall, exporting the encoder output as top CTC-trained logits introduces multiple benefits. The encoding achieves strong WER performance and interchangability is demonstrated through the modularity test. If regarded as a representation for first-pass ASR prediction, the Lego-Features surpass the N-best in quality, modularity, and generation speed.

To address resource-limited environments like on-device ASR, and to improve latency, future research can explore using smaller Exporter and Importer layers. Another avenue is to export CTC logits over phoneme/triphone/grapheme vocabularies, or a combination thereof. Different types of Lego-Features can be tested with various downstream tasks, like confidence models, speech translation or spoken language understanding.